# Understanding Social Perception, Interactions, and Safety Aspects of Sidewalk Delivery Robots Using Sentiment Analysis

Yuchen Du[a], Tho V. Le[a,*]

[a]*School of Engineering Technology, Purdue University, West Lafayette, IN 47907, USA*

**Abstract**

This article presents a comprehensive sentiment analysis (SA) of comments on YouTube videos related to Sidewalk Delivery Robots (SDRs). We manually annotated the collected YouTube comments with three sentiment labels: negative (0), positive (1), and neutral (2). We then constructed models for text sentiment classification and tested the models' performance on both binary and ternary classification tasks in terms of accuracy, precision, recall, and F1 score. Our results indicate that, in binary classification tasks, the Support Vector Machine (SVM) model using Term Frequency–Inverse Document Frequency (TF-IDF) and N-gram get the highest accuracy. In ternary classification tasks, the model using Bidirectional Encoder Representations from Transformers (BERT), Long Short-Term Memory Networks (LSTM) and Gated Recurrent Unit (GRU) significantly outperforms other machine learning models, achieving an accuracy, precision, recall, and F1 score of 0.78. Additionally, we employ the Latent Dirichlet Allocation model to generate 10 topics from the comments to explore the public's underlying views on SDRs. Drawing from these findings, we propose targeted recommendations for shaping future policies concerning SDRs. This work provides valuable insights for stakeholders in the SDR sector regarding social perception, interaction, and safety.

[*]Corresponding author. Email: thovle@purdue.edu

# 1. Introduction

With the advancement in economics, demographics, and information technology, the e-commerce sector has experienced significant growth. Retailers, aiming to establish brand identity and enhance customer satisfaction, have introduced more flexible options for parcel delivery in terms of timing and locations. This diversification in customer demands and the complexity of delivery tasks, along with a surge in freight volume, have made last-mile logistics increasingly intricate and challenging (1; 2). The reliance on traditional vehicular delivery has led to numerous issues, including escalated emissions, noise pollution, traffic congestion, elevated demand for parking spaces, and disruptions due to road maintenance or traffic incidents (1). Recent literature has discussed alternative transportation methods (3), such as robots, drones, cargo bicycles, electric vehicles, and their combinations, to address these challenges and promote sustainable transportation. However, these alternatives also present their own set of challenges: drones pose low payload, safety, and privacy risks, and cargo bicycles still require manual operation, demanding more labor (4; 2; 5).

Among the emerging technologies, Sidewalk Delivery Robots (SDRs) have increasingly gained prominence in last-mile delivery. These robots operate on sidewalks, sharing the space with pedestrians. Presently, they are predominantly utilized for food delivery in areas with lower population density or reduced traffic. Notable companies in this field include Starship [2], whose robots weigh less than 20 kilograms, can travel at a maximum speed of 6 kilometers per hour, and have a delivery range of approximately 3 to 5 kilometers. Each delivery can carry up to 10 kilograms of goods, with a cost of less than 1 Euro. These vehicles are equipped with multiple sensors, cameras, GPS, and other devices (6). As an innovative delivery tool, the introduction of SDRs has raised several new challenges. These include public unfamiliarity and apprehension, the lack of legal and regulatory mechanisms to govern these delivery robots, and the potential safety risks they pose to urban traffic. Researchers have begun examining existing legal and regulatory frameworks applicable to SDRs, identifying which could effectively supervise their operations (6; 7). Additionally, companies like Kiwibot are collaborating with the Pittsburgh government to study interac-tions between SDRs, pedestrians, and residents, further exploring the integration of these robots into the urban environment (8).

Our research is particularly focused on the public's perception of SDRs and the dynamics of their interactions with pedestrians. Several studies have already begun to explore this domain. In a pilot study of SDRs conducted in Pittsburgh, Pennsylvania, diverse perceptions and interactions of residents with these robots were observed. People with limited knowledge about the robots tended to speculate about their purposes and functionalities, while the robots' interactions with pedestrians, including children and pets, could cause distractions and obstructions. There were also instances where people assisted robots that were unable to move, highlighting potential accessibility issues (8). Different age groups have different attitudes towards SDR. The elderly group shows a tendency to refuse the introduction of SDR in urban public spaces. At the same time, different urban spaces also show different acceptance levels for SDR. While SDRs are widely accepted in public-controlled areas such as universities, their acceptance significantly decreases in areas with high population densities, such as city centers (9).

Moreover, for individuals with mobility impairments, the operation of SDRs on public sidewalks presents new challenges. They already face obstacles such as inadequate ramps and inconsiderate public behavior, and the addition of SDRs could further exacerbate these difficulties. This underscores the urgency of examining how SDR designs impact the daily navigation experiences of people with mobility impairments (10), and also, how SDRs affect the accessibility of public spaces (11). Additionally, a study assessing pedestrians' perceptions of sidewalk facilities revealed that safety is the paramount concern for pedestrians, regardless of their travel purpose. This emphasizes the importance of prioritizing pedestrian safety and comfort in the deployment of SDRs (12). In another study, researchers recorded video from ten locations on campus to understand the prevalence and severity of SDR-related interactions. Using the duration of encroachment to assess interaction severity, this study pinpointed predictors of both moderate and serious conflicts involving SDRs and pedestrians. (13).

Some researchers conducted questionnaire surveys to investigate public attitudes towards

SDRs. This study (14) examined the factors influencing Iranian online shoppers' adoption of SDRs. In their questionnaire, they incorporated considerations about environmental concerns, risks associated with using robots, factors of robot-human interaction, as well as several control variables (age, gender, education, income, and familiarity with delivery robots) to assess their impact on individuals' intentions. This study surveyed 287 participants. Another study (15) investigated consumers' willingness to pay for SDR services during the COVID-19 pandemic. By surveying 483 respondents, this study elucidated the potential consumer willingness to engage with SDRs across different consumer segments.

While questionnaires, videos and other forms of fieldwork can provide detailed information, they are often limited in scale and can be time-consuming to design and conduct. In contrast, data mining techniques can be used to gather and process large volumes of data automatically. The techniques are powerful tools to analyze public acceptance on a larger scale and are commonly used to provide ground truths and complement social experiment studies (e.g., questionnaire survey, social experiment, focus group interview). Social media platforms like Twitter and YouTube serve as rich data sources for such analysis. Researchers have used these data mining techniques and data from social media platforms to study topics related to transportation (16; 17). While there is a growing body of literature employing these methods to understand public perceptions in various domains, the application of such methods, specifically to the domain of SDRs appears to be less explored. This research contributes to the field by utilizing data mining techniques to analyze YouTube videos and comments, thereby offering a broader understanding of public perceptions related to SDRs that complement social experiment studies. In doing so, we aim to enrich the existing research and provide insights into (i) the prevailing sentiments about the use of SDRs, (ii) the key aspects of people's interactions and behavior with SDRs, and (iii) the pedestrian safety concerns associated with SDR operation. We perform sentiment classification and topic models to analyze the comments and suggest policies that could potentially benefit SDR management and usage in the future.

The remainder of this paper is organized as follows: Section 2 reviews the literature on relevant topics. Section 3 proposes a research framework and introduces the methodology, including Text preprocessing, details of Sentiment Classification, and Topic modeling. Section 4 introduces the data collection and labeling rules for sentiment classification. Section 5 presents the results and insights of our analysis. Section 6 discusses the caveats. Section 7 proposes potential policies related to SDR generated from the insights from the previous sections. And section 8 concludes this study with future research directions.

## 2. Literature review

### 2.1. Sentiment Classification

Sentiment classification (SC) is to classify texts according to the polarity of text data that corresponds to a label. In binary classification tasks, the labels are usually "positive" and "negative", and a ternary classification task would add a 'neutral' category. The general process of a SC task is shown in **Fig.** 1.

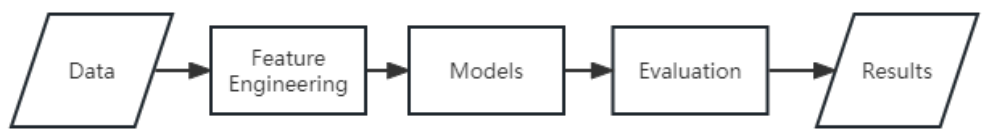

Figure 1: Sentiment Classification Process

*Feature Engineering* would significantly affect SC results. The most commonly used methods are Term Frequency–Inverse Document Frequency (TF-IDF) and N-gram for SC related to YouTube comments (18). The main goal of this step is to extract useful information from the raw text data to improve the performance of models.

The negation words 'not' and 'no' are important in SC since they can reverse the sentiment polarity of the sentence (19). When using the N-gram, different N values will greatly impact the results (20). (21) employed more common synonyms to replace infrequently used words and established a lexicon that only includes words with a frequency greater than number N. The purpose of this approach is to reduce the dimensionality. Furthermore, the dimension of the word vector is determined by the mean text length plus its standard deviation. (22) performed part-of-speech tagging on all words and appended the POS tag as a suffix to each word in the form like 'word POS'. Then they selected nouns, verbs, adjectives, adverbs, and prepositions as the inputs. Additionally, they paid attention to the negation words "not" and "no", combining bi-gram phrases such as "not word" into the form "not_word" with "not" and "no" as prefixes.

Deep learning (DL) models can be used to extract richer feature information from the text and achieve better performance in classification tasks. Researchers can capture the similarity between words by converting words into vector representations using a method called Word Embeddings, including Word2vec (23), Glove (24), and Fashtext (25). The Bidirectional Encoder Representations from Transformers (BERT) (26) is another powerful method for text representation. Unlike traditional word embeddings, BERT takes into account the full context of a word by looking at the words that come before and after it, making it particularly effective for understanding the semantic meaning of words and sentences.

*Models* commonly used in the SC of YouTube comments are Naive Bayesian (NB), Support Vector Machine (SVM), K-Nearest Neighbors(KNN), and Decision Tree (DT) (18). (27) used Naïve Bayes – Support Vector Machine (NBSVM) Classifier to perform binary SC task on YouTube comments. First, they employed the Multinomial Naive Bayes (MNB) method to compute the probability of each word's appearance under positive and negative labels. They transformed these probability values into feature vectors to represent the words. Then, these feature vectors served as the input for the SVM to get the hyperplane utilized for the classification. They got a precision of 0.91 and F1 score of 0.87 on 333 comments. (28) tested SVM, KNN, and Bernoulli Naïve Bayes on Arabic comments from Youtube. It is noteworthy that the dataset utilized in this study was highly imbalanced. Out of a total of 5,986 data points, 4132 were positive, 780 neutral, and 1,074 negative. The researchers generated balanced and imbalanced datasets from this original dataset and used these for binary classification (containing only positive and negative labels) and ternary classification. The findings indicate that applying the SVM to the imbalanced binary dataset yielded optimal results, achieving a precision and recall rate of 0.89. (29) used the NB for SC of movie trailer reviews on YouTube. The results of the study show that NB achieved an accuracy rate of 0.81, a precision rate of 0.75, and a recall rate of 0.75. (30) compared the performance of NB, SVM, and KNN for SC. The results show that the average accuracy of SVM is 0.96, which is the best among all models.

As for DL models, Long Short-Term Memory Networks (LSTM) and Gated Recurrent Unit (GRU) are commonly used in SC. Both are improved versions of the Recurrent Neural Network (RNN). LSTM and GRU can effectively process text sequence data and solve the problems of gradient disappearance and gradient explosion in RNN (31). (32) compared Deep Neural Networks (DNN), Convolutional Neural Networks (CNN), and LSTM on different datasets and found LSTM outperformed other models. Some other applications on YouTube or Twitter, such as (33) used word2vec and SVM, (34) tested Glove and Fashtext with LSTM, and (21) used word2vec and CNN.

*Evaluation* metrics are used to compare the performance of different models. In SC, we often use Accuracy, Precision, Recall rate, and F1 score.

### 2.2. Labeling

Formulating the annotation rules according to the research goals is necessary before starting labeling. Data labeling usually requires manual processing, but in some cases, we can use information such as rating scores from Google Store to label data automatically (35). For the types of sentiment labels, in simple tasks, we usually only need to distinguish whether the sentiment expressed by the text is negative, positive, or neutral, but for more complex SC tasks, the text needs a more complete sentiment label to describe. One way is to build a more complete emotional model, such as Ekman's Six Basic Emotions Model (36) and Plutchik's Wheel of Emotions (37; 38). The other way is to define labels according to the subjectivity and emotional strength of the text (39), such as " strongly negative", "negative", "neutral", "positive", and "strongly positive" (40).

### 2.3. Topic Model

Topic modeling is a type of statistical model used for discovering the abstract "topics" that occur in a collection of documents. It can be used to identify patterns in a corpus. The topic model assumes that each document is generated by mixing multiple topics, and each topic can be regarded as a probability distribution of words. The Latent Dirichlet Allocation (LDA) model (41) is the most popular and most studied model in many fields (42). Some researchers have used the LDA model to analyze traffic-related topics on social media, such as Twitter (17).

## 3. Methodology

*3.1. Research Framework*

This research will use machine learning (ML) and DL methods and models to analyze the comments from YouTube videos to fulfill the research goals we proposed in the Introduction section. The main steps involved in this study are shown in **Fig.** 2.

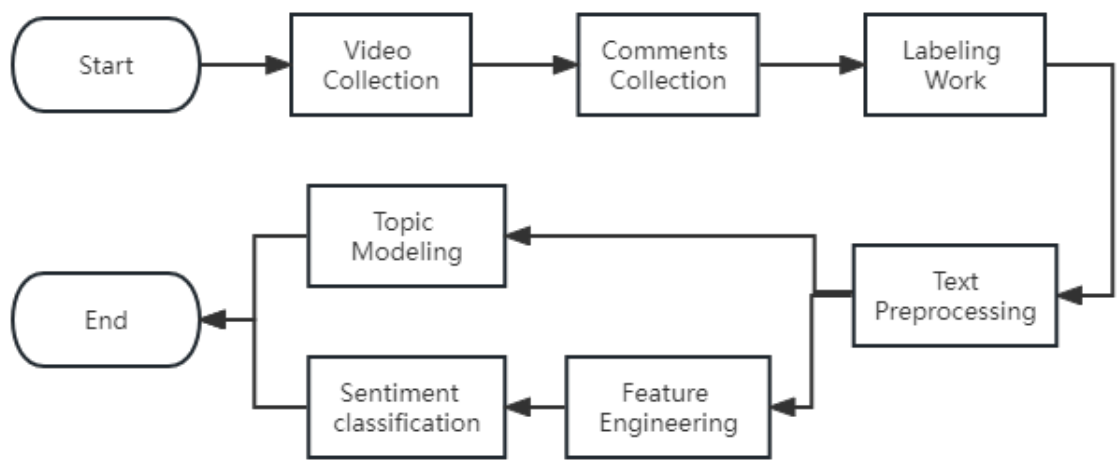

Figure 2: Research Framework

1. **Video Collection:** Search a list of videos in English on YouTube using keywords for sidewalk delivery robots and save these videos on the playlists.

2. **Comments Collection:** Use the YouTube API to extract the comments of the videos we saved.

3. **Labeling work:** Formulate labeling rules and label the comments we collected.

4. **Text preprocessing:** Prepare and clean the comments.

5. **Feature Engineering :** Transform the comments into numerical vectors or representations that can be understood and used by ML models.

6. **Sentiment Classification:** Build ML models for this SC task, thereby obtaining the model performance and categorization outcomes.

7. **Topic Modeling:** Use the topic model to generate topics from the comments to find underlying aspects of the comments.

The following sub-sections introduce the techniques and methods used in this research regarding text preprocessing, feature engineering, modeling, evaluation, and topic modeling. For the ML part, We used NLTK (43) to implement the text preprocessing and used Scikit-learn (44) to implement feature engineering and build models, including MNB, SVM, and DT. For the DL model, we used pre-trained BERT from HuggingFace (45) and LSTM, GRU from Pytorch (46). There are two notes to make here: (1) We only used the DL model in the ternary classification task; (2) For the DL model, we employed BERT for text representation. As a pre-trained model, BERT already includes text preprocessing and feature engineering. Consequently, in the following Text processing and Feature engineering sections, we only focus on the methods used in our ML models.

*3.2. Text Preprocessing*

The text processing steps are shown in **Fig.** 3.

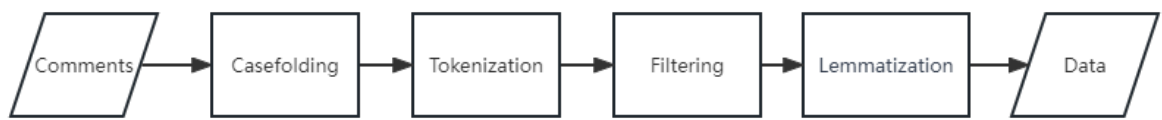

Figure 3: Text Preprocessing

The four steps convert text comments into data: *Casefolding* - is a process to convert all text to lowercase; *Tokenization* - is the process of decomposing text into individual words or symbols. These individual words or symbols are called "tokens". For example, the sentence "I like SDRs." will be decomposed into "I", "like", "SDRs" after tokenization; *Filtering* - is a process to remove the stopwords and some special characters such as "@" and "#"; *Lemmatization* - is the process of converting words into their base form (or root form), which helps extract the text's main meaning. For example, "running", "runs", "ran" all become their root "run" after lemmatization.

*3.3. Sentiment Classification*

We organize this section following the steps shown in **Fig.** 1.

*3.3.1. Feature Engineering*

In this research, we focused on punctuation, negation words, low-frequency words, and N-gram & TFIDF to perform feature engineering on the dataset.

Punctuation: The punctuations here are ["?", "!"]. We preserved these two punctuation marks in the text. In our dataset, we noticed that exclamation marks "!" are often accompanied by strongly emotional words. In contrast, question marks "?" are concentrated in comments expressing people's confusion about SDRs, which are usually related to the functions, advantages, or disadvantages of SDRs. Therefore, we believe that these two punctuation marks can provide useful information to the model.

Negation words: We preserved the negation words "not" and "no" because these two words could have a big impact on the sentiment of a sentence.

Low-frequency words: We removed low-frequency words, which are defined in this paper as words that appear only once across all comments. When creating a vocabulary and vectorizing words, removing low-frequency words can effectively reduce the dimension of word vectors and speed up model training, which is especially effective for some simple models (such as NB), but with the trade-off cost of losing some information.

N-gram & TF-IDF: We used N-gram to create more features and Term Frequency-Inverse Document Frequency (TF-IDF) to change our text data into vectors. N-gram is a method based on a statistical language model, which performs a sliding window operation of size 'N' on the content in the text to form a continuous sequence of words. For example, when we use 2-gram, the sentence 'I like SDRs' will be changed to ('I like', 'like SDRs'). TF-IDF is another feature engineering technique of representing text, which can be used to transform unstructured text data into structured numerical data (47). The TF-IDF formula for calculating word $i$ is as follows:

$$\text{TF-IDF}(i) = \text{TF}(i) \times \text{IDF}(i) \tag{1}$$

$$\text{TF}(i) = \frac{\text{the number of times word } i \text{ appears in the document}}{\text{the total number of words in the document}} \tag{2}$$

$$\text{IDF}(i) = \log_e \frac{\text{total number of documents}}{\text{number of documents with word } i} \tag{3}$$

*3.3.2. Models*

For ML models, we implemented MNB, SVM with RBF kernel function, and DT with default parameters. We tested the task for both binary and ternary classification. The binary classification was subdivided into six subsets according to the labels: 0-1, 1-2, 0-2, 0 versus others, 1 versus others, and 2 versus others. Within each classification task, we examined the optimal value of 'N' in the N-gram that achieved the highest accuracy for different models.

For the DL model, we used Bert, LSTM layer, and GRU layer. We only used this model in ternary classification tasks compared with other ML models. After using BERT to tokenize the comments, 75% of the text lengths are within 120 tokens. Therefore, we set 120 as the maximum sequence length for input into the model. When encountering sequences that exceed this length, the sequence will be truncated, retaining only the first 120 tokens. The architecture details are shown below:

Model Architecture:
1. BERT (Bidirectional Encoder Representations from Transformers)
   - Pretrained model name: BERT base model (uncased)
2. LSTM (Long Short-Term Memory)

- Hidden size: 512
- Bidirectional: True
3. GRU (Gated Recurrent Unit)
    - Input size: 1024 (output from LSTM layer)
    - Hidden size: 512
    - Bidirectional: True
4. Dropout Layer
5. Linear Layer (Intent Classifier)
    - Input size: 1,024
    - Output size: 3 (number of labels)

*3.3.3. Evaluation*

Since it is a multi-class classification problem, we used weighted average values of accuracy, precision, recall rate, and F1 score to evaluate the performance of the models. The weights were obtained by the proportion of the three types of comments denoted as labels 0, 1, and 2 in the test set.

To calculate these performance metrics, we need to know the confusion matrix first. This matrix is shown in **Fig.** 4. In this matrix, TP means the instances that the model correctly identified as positive (or "true"), and FP means the instances that the model incorrectly identified as positive when in fact they were negative. TN and FN have similar meanings.

Then we compute accuracy, precision, recall rate, and F1 score.

$$\text{Accuracy} = \frac{TP + TN}{TP + FP + TN + FN} \quad (4)$$

$$\text{Precision} = \frac{TP}{TP + FP} \quad (5)$$

$$\text{Recall} = \frac{TP}{TP + FN} \quad (6)$$

$$\text{F1 socre} = \frac{2 \times \text{Precision} \times \text{Recall}}{\text{Precision} + \text{Recall}} \quad (7)$$

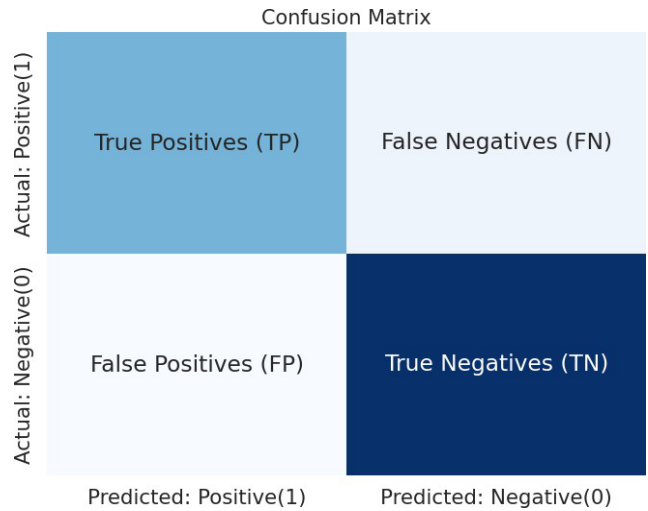

Figure 4: Confusion Matrix

*3.4. Topic Model*

We used Gensim (48) to implement the LDA model and used Confirmation Measure (49) to measure the quality of the topics generated by the model. This metric is a measure based on word pairs, considering the similarity between words and the frequency of words in the entire corpus, reflecting whether the words in the topic are consistent. We want to find topics that are rich in content and have high consistency scores to better study the content of comments in the dataset. Higher Confirmation Measure scores indicate better quality of generated topics.

**4. Data Collection and Comments Labeling**

We first used keywords related to SDRs to search the videos on YouTube. These keywords are "sidewalk delivery robot", or "delivery robot", or "autonomous vehicles", or "SDR". We endeavored to exhaustively search for all keywords pertinent to sidewalk delivery robots, thus employing terms akin to 'SDR' such as 'autonomous vehicles'. Subsequently, appropriate videos were selected through a manual filtering process.

The YouTube platform currently has two types: Short and regular videos. The distinction between the two types of videos is the duration, with the former typically lasting less than one minute. Then We archived these two types of videos in 3 distinct playlists based on the number of comments, named "SDR 10 50", "SDR 50 100", "SDR 100 1000". For example, "SDR 10 50" stored all videos with 10 to 50 comments. We did not include videos having less than 10 comments because they can not provide rich information. The details about this step are shown in **Fig.** 5, where N represents the number of videos.

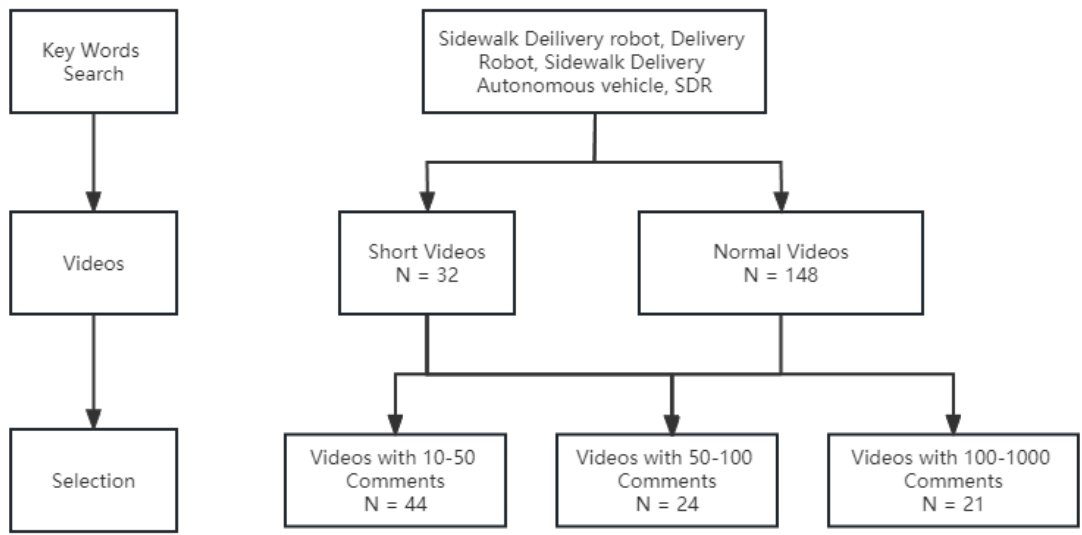

Figure 5: Video Collection

After collecting the videos, we used YouTube API to extract the comments of each video in the playlists. Then we moved to the *labeling work* step. In this study, we have categorized the labels into 3 categories: negative (0), positive (1), and neutral (2). The annotation rules are formulated around the three research goals we mentioned in the introduction section. We employed a manual annotation method, where two independent annotators labeled the comments. The final version of the annotations was determined through consultation and discussion between the two individuals. The details of the rules are shown below:

- Label **0**: We label comments that display hostility, disgust, or aggressive intent towards SDRs, criticisms of the shortcomings of SDRs, and concerns about the potential negative impacts of SDRs as negative comments.

- Label **1**: We label comments that show love, support, praise, or protective tendencies towards SDRs, discussions on the advantages of SDRs, and exhibit an optimistic attitude towards this technology as positive.

- Label **2**: Comments that do not exhibit a clear emotional tendency.

**Label statistics**: After labeling the comments, we obtained a total of 4,971 comments pertaining to SDRs. Among these, 2,101 comments (42%) were labeled as '0', 1,228 comments (25%) were labeled as '1', and 1,642 comments (33%) were labeled as '2'.

## 5. Results and Discussion

### 5.1. Text Preprocessing Results

**Table.** 1 shows the results of each text preprocessing step. As mentioned earlier, we keep 'not', 'no', ' !', and '?' since the first two words can flipper the sentiment of the sentence, while ' !' shows in the sentence with the strong sentiment, ' ?' shows in the ones with confusion, which are useful information for our SC task.

### 5.2. Prediction Results

The results are listed in **Table.** 2. To save space, only the model with the best accuracy is listed for each binary classification task. In column 'Model', '+ (1,3) gram' means we considered 1-gram, 2-gram, and 3-gram at the same time. This implies that we used single words, consecutive two words, and consecutive three words as features.

The results show that the value of N has a significant impact. For the ML approach, SVM performs the best on most binary classification tasks, while MNB performs the best on three classification tasks. It can also be seen from the table that the ML model has the strongest ability to recognize label 1 but is confused when classifying between labels 0 and 2. Part of the reason may come from our definition of labels 0 and 2, resulting in many similar words or phrases in these two types of comments. Another finding is that binary classification results are always better than ternary classification results.

In the context of this research, the performance of the DL model is significantly better than all of the ML models we used, and an accuracy rate close to 0.80 can be achieved.

| Step | Results |
|---|---|
| Youtube comments | I'm not sure about the robot. It seems a bit slow and could potentially block the sidewalk. |
| Casefolding | i'm not sure about the robot. it seems a bit slow and could potentially block the sidewalk. |
| Tokenization | ['i', "'m", 'not', 'sure', 'about', 'the', 'robot', '.', 'it', 'seems', 'a', 'bit', 'slow', 'and', 'could', 'potentially', 'block', 'the', 'sidewalk', '.'] |
| Filtering | ['not', 'sure', 'robot', 'seems', 'bit', 'slow', 'could', 'potentially', 'block', 'sidewalk'] |
| Stemming | ['not', 'sure', 'robot', 'seem', 'bit', 'slow', 'could', 'potential', 'block', 'sidewalk'] |

Table 1: Example results of text preprocessing

## 5.3. Topic Model

To generate high-quality topics, we removed certain frequently occurring words that do not provide valuable information after a series of tests. Initially, these were interjections ('ha', 'oh', 'yeah', 'wow', 'lol') and certain adverbs ('also', 'really', 'actually'). Subsequently, we removed the words 'robot', 'delivery', and 'human' since these words appeared in all topics and did not contribute beneficially to differentiating among the topics. **Fig.** 6 showed the results of the Confirmation Measure (CM) under different numbers of topics.

According to the peak value in the figure, we selected the topic contents when the number of topics was 5, 6, 7, and 10 for comparative analysis. Although the CM value with 10 topics was slightly lower than those of CM values with 5, 6, and 7 topics, it provided wider and richer content for deeper analysis, so we finally decided to present 10 topics which are shown in **Fig.** 7. The horizontal axis represents keywords under each topic, and the vertical axis represents the probability of occurrence of keywords under the topic.

| Model | Label | Accuracy | Precision | Recall | F1 |
|---|---|---|---|---|---|
| **MNB + (1,2) gram** | 0,1 | 0.84 | 0.84 | 0.84 | 0.83 |
| **SVM + 1 gram** | 0,2 | 0.76 | 0.75 | 0.76 | 0.76 |
| **SVM + 1 gram** | 1,2 | 0.79 | 0.79 | 0.79 | 0.78 |
| **SVM + (1,4) gram** | 0, other | 0.77 | 0.77 | 0.77 | 0.76 |
| **SVM + (1,2) gram** | 1, other | 0.86 | 0.86 | 0.86 | 0.84 |
| **SVM + 1 gram** | 2, other | 0.74 | 0.74 | 0.74 | 0.70 |
| **MNB + (1,3) gram** | 0,1,2 | 0.68 | 0.70 | 0.68 | 0.68 |
| **SVM + (1,2) gram** | 0,1,2 | 0.69 | 0.70 | 0.69 | 0.68 |
| **DT + (1, 2) gram** | 0,1,2 | 0.60 | 0.60 | 0.60 | 0.60 |
| **BERT + LSTM + GRU** | 0,1,2 | 0.78 | 0.78 | 0.77 | 0.78 |

Table 2: Prediction results

In **Fig.** 7, frequently occurring keywords reveal underlying aspects of the comments. We investigate all topics to reveal the perception, interaction, and safety aspects regarding DSRs. We use comments from the dataset to help better explain the meaning of topics, therefore, we preserve the nature of the comments without any modifications on texts, capital letters, etc.

Topic 1 has a mix of positive ('cute', 'cool', 'need', 'love') and negative words ('job', 'take', 'steal', 'replace'). It reveals YouTube users' concerns about robots replacing humans and taking away human jobs. Topic 1 is then labeled as Job security concerns.

> ***Takes*** *away a lot of **jobs***
> *This is totally gonna **steal jobs**. Even if it is, for now, piloted by a people.*
> *Wake up people you are starting to be **replaced!!***
> *Another **job** that 'eventually' will be completely **replaced** by a machine.*

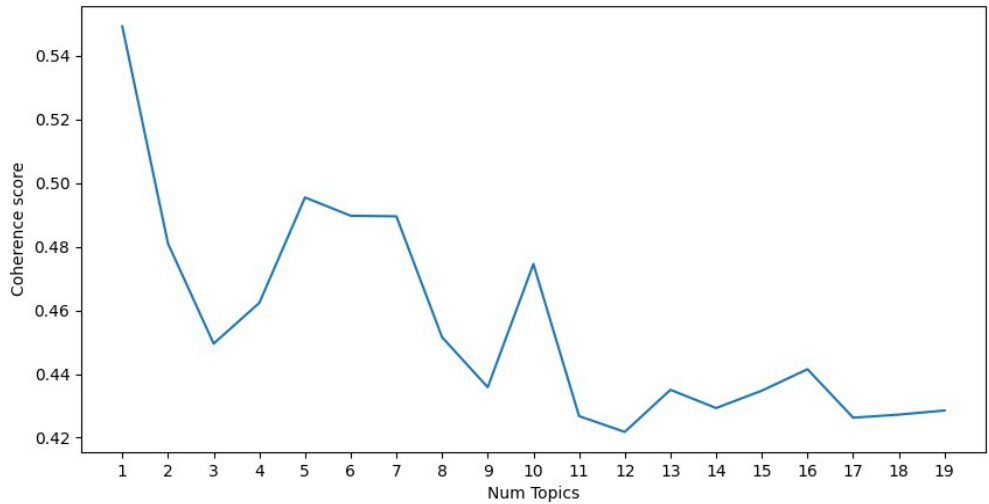

Figure 6: Confirmation Measure scores as a function of the number of topics

Topic 2 refers to the main usage of SDRs for food delivery ('order', 'food'). Topic 2 is then labeled as Foods delivery applications.

> *Imagine just **ordering food** and then you just see that coming towards you*
> *Hopefully it knows if you are the real person that **ordered** something. Not a thief.*
> *Awesome idea. You still have a person pick the **food** up, so idk why he can't just continue like other apps. I understand he can grab multiple **orders**, but with that won't the first **order** get cold?*
> *when the **food** you **order** is slower than people walking*

Topic 3 shows the ideas of the 'future' of SDR technologies. Topic 3 is labeled as Technology of the future.

> *In this **future** no one **works**. we are all home ordering food and usb cables.*
> *What is this world **coming** to? I personally do not think this is **good** for our future...*
> *I like the way our future is rapidly expanding and **coming** super intelligent when it **comes** to helping mankind*
> *it just makes me feel like the **future** is **coming***

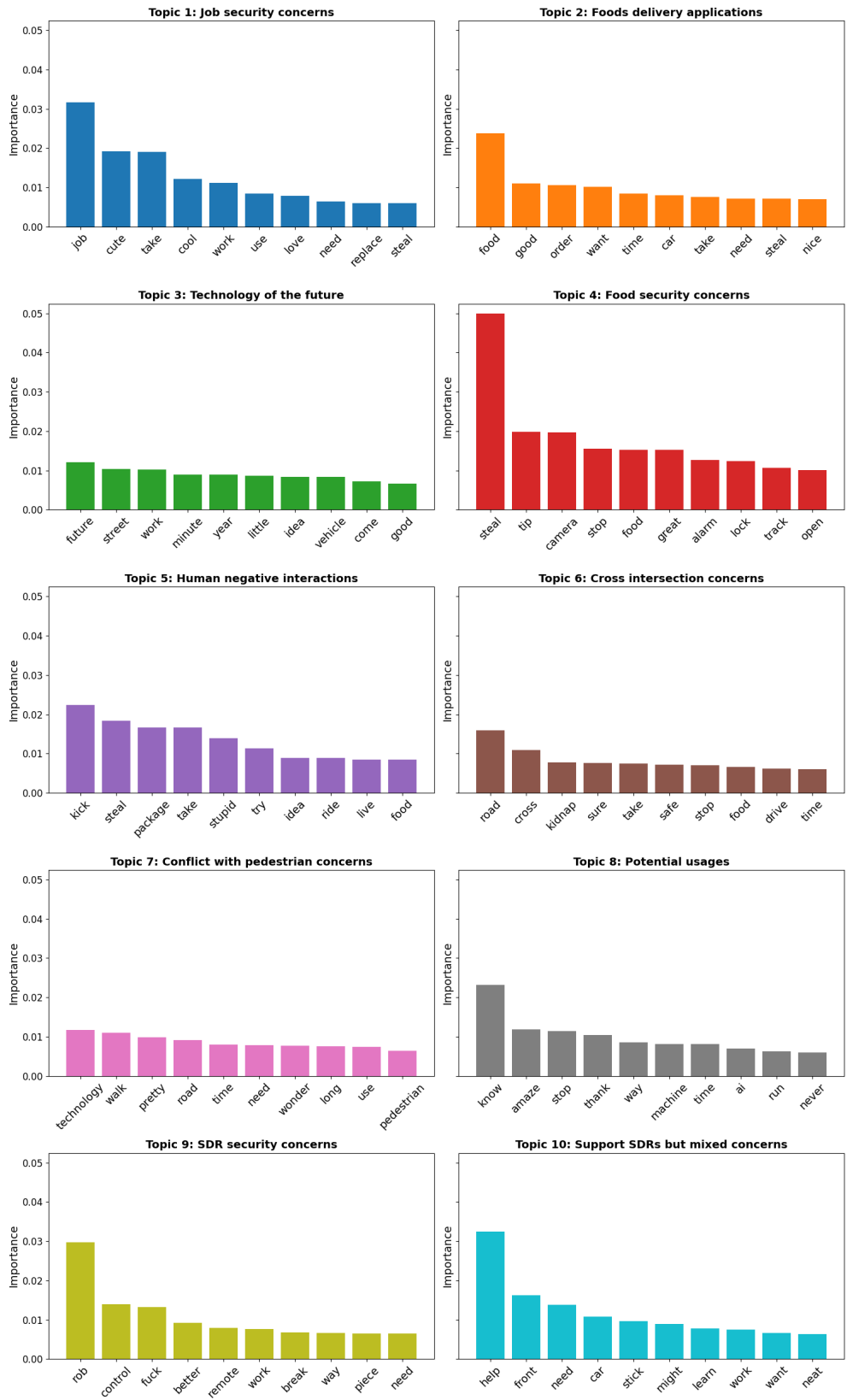

Figure 7: Keywords distribution of 10 topics

Then come to the negative topics 4, 5, 6, and 9. Topic 4 shows potential security issues of SDRs ('steal', 'tip', 'stop', 'open') and the anti-theft measures of SDRs ('alarm', 'lock', 'camera', 'track'). Topic 4 is labeled as Food security concerns.

> *Unless someone on the streets **stops** it and **steals** our food.*
> *Thieves and cost will **stop** them.*
> *Roll it on its side. Or **TIP** IT OVER!*
> *Some hooligan is just gonna break it **open** and **steal** the package.*
> ***Alarms**, **Cameras**, Robots would be destroyed, food **stolen** in NYC, even Atlanta Georgia.*

Topic 5 also presents some negative actions ('kick', 'take', 'steal', 'ride') around the objects ('package', 'food'). Some comments mentioned this robot is a 'stupid idea'. Topic 5 is labeled as Human negative interactions.

> *Hey here comes my free **ride** downtown.*
> *Not many estates in the UK will be a safe place for these, and it is getting less safe!! Has anyone ever seen a child sitting on one for a **ride**?*
> *Just **kick** him and destroy it.*
> *I would just **kick** the robot over and **steal** the food.*

Topic 6 and 9 present new safety concerns ('kidnap', 'rob', 'break'), new scenarios ('cross', 'road', 'drive', 'time'), and new operating functions ('control', 'remote'). Topics 6 and 9 are labeled as Cross intersection concerns and SDR security concerns, respectively.

> *That robotic delivery Rover stuck in **road cross**, because of traffic.*
> *What happens when the robot falls sideways in the middle of the **road**?*
> *So will these robots get tickets for jaywalking? I noticed one **crossing** but not at a crosswalk.*
> *It's all fun and games until that thing gets hit by a car while **crossing** the street.*
> *Easily can **kidnap** robot throw it in car trunk*
> ***road** hazards. should be banned.*

> *It's illegal to **drive** on the sidewalk!*
>
> *So one pack at a **time**? this seems very inefficient.*
>
> *THE ROWDY KIDS ARE gonna DESTROY & **ROB** THOSE THINGS.*
>
> *How it being **controlled***
>
> *There is actually someone piloting that via **remote control***

Topic 7 presents concerns on potential conflicts of SDRs with 'pedestrians'. Related statements centered on the comments that robots should operate on 'roads' rather than sidewalks. This topic also includes content related to 'technology' and 'time'. Topic 7 is labeled as Conflict with pedestrian concerns.

> *Look how much space on the sidewalk that thing takes up! I wouldn't be a happy **pedestrian**. This thing would be annoying.*
>
> *Motorized vehicles belong in the **road** not on the sidewalk/ pavement where they are a hazard to **pedestrians** and people with disabilities.*
>
> *These robots are a public nuisance they are very intimidating for the elderly and **pedestrians**.*
>
> *It fills all **walk road**?*
>
> *Pavements are for people. These things should be on the **road**.*

Topic 8 shows some of the people's understanding ('know', 'amaze', 'thank') of robots and content about 'ai' and 'machine'. Topic 8 is labeled as Potential usages.

> *Very cool. I have seen these robots before but didn't **know** anything about them.*
>
> *This will be **amazing** for the elderly. I **never know** this existed. get this in buffalo Ny ASAP. there are senior homes people can use this.*
>
> *anyone **know** how much it would cost to buy it?*
>
> *Yeah I am really afraid of **AI** taking over*
>
> *This is how Skynet and the Future **Machine** Wars starts*

Topic 10 shows people's mixed attitudes when faced with whether to 'help' a robot, robot-'car' interaction, and robot effectiveness ('might', 'work', 'need', 'front'). Topic 10 is labeled as Support SDRs but mixed concerns.

*happy to **help** the tiny robot.*

*Don't **help** robots take our jobs. This is dumb as fk.*

*That small flag is supposed to make the robot visible to **cars** and **pedestrians**?!?! I think it has to be at least as noticeable as humans to avoid causing collisions!*

*I would get in **front** of it and walk as slowly as possible, or drop a large box over it so that it couldn't navigate.*

*Seems like a **neat** idea for last mile delivery. **Might need** a bit of self-defense just in case.*

*5.4. Insights*

According to the SC prediction results and the topic model results, we now discuss the three research questions that we proposed at the beginning of this paper.

**(i). What are the people's prevailing sentiments about the use of SDRs?**

Based on the fact that approximately 40% of our dataset's comments are negative, coupled with our subsequent analysis, YouTube users generally have a doubt or negative attitude toward SDRs. People's concerns mainly come from robots replacing humans, taking away human jobs, reducing wages, food security concerns, safety concerns, etc. Many delivery drivers have commented on the videos expressing their concerns, especially when SDRs do not require customer tipping. Some comments even worry that AI will take over the future society. For example, "Skynet" appears repeatedly in our dataset. This doubt and negative attitude reflect the human commonsense of deficiency needs as captured in Maslow's hierarchy of needs. The basic survival concerns are physiological needs and safety needs. In which, personal security, employment, health, and property are of high-level in need. Topics 1, 4, 6, 7, 9, and 10 confirm the findings.

About a quarter and a third of the comments reflect the positive and neutral attitude toward the use of SDRs. The attitude is confirmed by topics 2, 3, 8, and 10. People think SDRs are the technology of the future that can be used for multiple applications including food delivery in urban and small areas that serve people and elderly groups, etc.

**(ii). What are the key aspects of people's interactions and behavior with SDRs?**

People are mainly concerned about vandalism against robots, such as breaking the lid of the robot to steal the food inside, kicking the robot over, or kidnapping the robot and taking it away with a car. Although these robots already have certain anti-theft measures, such as GPS, trackers, facial recognition and alarm, destroying a small robot is still an easy job judging from the videos we collected. Topics 5, 9, and 10 reflect the concerns.

In addition to the content of the topics we generated, we were able to observe additional aspects from the videos and comments we collected: One is the food problem. Some videos show that the coffee ordered by the customer spills inside the robot, and the smell of the last delivered food remains inside the robot. Another issue is that during Covid-19, there are concerns that SDRs would spread disease, and people are concerned about whether the hands of shop assistants who put food into the robots are sanitized. Topic 4 provides more insights into this issue.

**(iii). What are the pedestrian safety concerns associated with SDR operation?**

The concerns in this regard mainly come from the fact that SDRs occupy the sidewalk space that belongs to pedestrians and bring safety hazards to the elderly, children, and visually-, hearing-, or mobility-impaired people. Some people in their comments questioned whether it would be legal for the robots to drive on sidewalks. At the same time, these robots also present a road safety hazard. A small SDR and a tiny flag could not be visible enough for drivers to spot them in the rearview mirror in time, and the robots may also break down and stop when crossing the road, or not finish the crosswalk within the given time, which could cause serious traffic accidents. Topics 6 and 7 provide deeper insights into these concerns.

## 6. Caveats

In this section, we discuss the caveats that could affect the performance of SA.

*6.1. Context Information & Video Characteristics*

The content of a YouTube video provides a specific context and emotional undertone to its comments. Therefore, similar comments under different videos could express contrasting sentiments, especially when people used irony or metaphor. Capturing this complex semantic information from a small dataset is a challenging task for modeling.

Moreover, the video content itself can significantly influence the sentiment of the comments. Poor choice of background music can provoke negative reactions from viewers, leading to negative comments. Some videos selected inappropriate interviewees, interviewing obese people's views on robot food delivery, which led to a large number of comments to the video that "robots will make humans lazy and obese", which is very seriously misleading.

Although the content of the videos themselves exerts a certain influence on the comments, the predominance of videos inclined to express negative sentiments within our collected comments suggests an indicative trend regarding current public attitudes towards SDRs. This inference is predicated on the fact that we have endeavored to gather the majority of high-viewership videos pertaining to SDRs available on the YouTube platform to the best of our ability.

*6.2. Misspelling*

This issue was also discussed in (22). Given the relatively small size of our dataset, which consists of around 5,000 comments, misspellings may create 'neologisms'. During the feature engineering phase, we removed words that appeared infrequently, which sometimes included these misspelled words. This step could potentially lead to some valuable information loss. For instance, if 'robot' is misspelled as 'robott' or 'robo', these words may be eliminated due to their low frequency. We may therefore be missing important information in these comments related to 'robot', resulting in less accurate SC.

*6.3. Data Quality*

Some comments may contain elements of irony or metaphor, which represent prevalent challenges within the field of Natural Language Processing (NLP). Our adoption of manual annotation has, to some extent, mitigated these issues through human discernment. Furthermore, unlike e-commerce platforms, which operate as a marketplace for selling goods, YouTube does not share the same strong incentives that might motivate individuals to fabricate comments.

*6.4. Labeling Errors*

No matter how advanced the models are used, defining clear annotation rules based on research goals is always the most important step. If the differences between the meanings of various labels are unclear, it could lead to similar comments being categorized into different labels. This could make the classification task harder. For example, (28) labeled texts including both positive and negative words as neutral. In subsequent research, researchers found that such labels posed serious challenges to classification. In our dataset, people's apprehensions about SDRs often touch on topics related to the robots' functionality. This results in many errors when the model classifies labels 0 and 2. Insufficient data volume is also a factor that affects this problem.

**7. Policy Considerations**

The aim of policy development is to ensure the economic benefits of SDRs are realized in a safe, controllable, and harmonious manner with humans, thereby enhancing the efficiency of last-mile delivery. However, formulating and implementing such policies is a complex task, necessitating the consideration of various stakeholders' interests. Governments need to balance the interests of companies by supporting innovation and development related to SDRs, while also considering the impact of SDRs on urban traffic and potential risks to the public. Policy considerations for SDRs are challenging due to their lack of widespread adoption. Fortunately, extensive policy experience exists with electric and autonomous vehicles, which are somewhat analogous to SDRs, though differences lie in the sidewalk operation of SDRs. Our research presents administrative considerations for different stakeholders.

*7.1. For Legislators and Governments:*

1. Legal considerations: Authorities need to define what constitutes an SDR, their technical standards, and liability assignment. This includes determining liability and compensation in traffic incidents involving SDRs.

2. Workforce skill development: Addressing concerns about robots replacing human jobs, governments should collaborate with companies to provide necessary training related to SDR operation, and jointly consider the roles of delivery drivers and SDRs in delivery services, exploring complementary rather than conflicting delivery models.

3. Industry support: Governments should assist in developing infrastructure for SDRs, similar to electric and autonomous vehicles. This includes charging stations, storage facilities, and offering incentives or support to local businesses utilizing SDRs. Additionally, establishing management agencies to address issues arising from SDR deployment may be necessary.

*7.2. For Urban Traffic Managers:*

1. Traffic control measures: Establish measures clarifying the behavior of SDRs under different scenarios, on various roads, intersections, and in different city areas.

2. Infrastructure support: Ensure separate pathways for SDRs and pedestrians, where possible. The future sidewalk and street design or maintenance should consider SDRs as special users of different transportation infrastructures.

*7.3. For Consumers' Perspective:*

1. Data security: Concerns about the security of private data, similar to drones. SDRs can collect geographic location information and delivery trajectories of users. Legislators and governments should carefully consider the acquisition and application of data collected by SDRs for transportation management purposes.

2. Customer terms development: Encourage SDR-operating companies to develop detailed customer terms, clarifying issues like liability and compensation for lost delivery items, thereby protecting consumer rights.

*7.4. For Companies' Perspective:*

1. Cooperation with governments: Companies should cooperate with governments in implementing measures for the safe operation of SDRs and share costs in certain areas.

2. Equitable access and design considerations: Provide subsidies in remote and low-income areas to make SDR services more affordable. The design of SDRs should consider the specific needs of different communities, such as offering multilingual support and being simple to use for those with mobility impairments or who are not tech-savvy.

*7.5. Other considerations:*

While the findings of this study revealed topics on different aspects of food delivery, safety, traffic management, and job implications related to SDRs, it is important to acknowledge that the policy considerations surrounding SDRs are extensive and multifaceted. Our current analysis, guided by the dataset at hand, provides focused insights into specific areas of public concern. However, several other dimensions warrant future exploration to enrich our understanding of SDR-related policy considerations.

For instance, aspects such as the charging infrastructure for SDRs, which plays a crucial role in their operational efficiency, have not been discovered by comments in this study which were mainly provided by the SDR service users, not the SDR operators. Additionally, issues related to equity in access to SDR services are vital for ensuring that these technologies benefit a broad spectrum of the population, avoiding the creation of a digital divide. These aspects, while not supported by our current dataset, are critical for the comprehensive development and implementation of SDR policies.

Future research efforts should aim to collect and analyze data relevant to these aforementioned dimensions. By expanding the scope of data collection to include such aspects, subsequent studies can offer a more holistic view of the policy landscape surrounding SDRs. This approach will not only enhance the depth of policy formulation but also ensure that it is inclusive and considerate of various stakeholder needs.

## 8. Conclusions

This study provides an in-depth SA of public attitudes toward SDRs based on comments from YouTube videos. Our survey results show a large negative sentiment toward SDR, with concerns centered around job losses due to the robots, vandalism and theft of robots, and the potential safety hazards robots may pose to pedestrians, cyclists, and cars. Our approach leverages the vast amount of data available in online comments compared to traditional research methods such as questionnaires with limited sample sizes. The use of ML and DL models enables the processing of large datasets, providing more comprehensive and ground-truth insights into public sentiment.

Also, we conducted SC tasks on comments labeled 0, 1, and 2, testing both binary and ternary classification tasks. Our results indicate that among the binary classification tasks, the SVM model achieves the highest accuracy of 0.86 in the '1 vs others' task. In the ternary classification task, the DL model combination of BERT, LSTM, and GRU significantly surpassed the ML models, reaching an accuracy of 0.78. These findings underscore the effectiveness of SVM and the potential of DL models in SC tasks.

This study, through an analysis of YouTube comments on SDRs, has revealed a diverse range of public attitudes and concerns towards this technology. The results of the topic modeling showed that public perceptions of SDRs encompassed aspects ranging from job security, food delivery applications, prospects of future technology, and concerns about food and robot safety, to potential conflicts between SDRs and pedestrians. While some people are excited about the technology and its usefulness, a few others show mixed feelings. A large portion of people are still in doubt or concerned about the negative impacts of SDRs. These findings not only enhance our understanding of the societal impacts of SDRs but also provide an empirical basis for policy formulation.

In light of the policy considerations, we have proposed a set of recommendations aimed at addressing the key issues identified in the topic model. These policies encompass enhancing job security, supporting local businesses, improving transportation management, and elevating the safety and security of SDRs. The implementation of these policies will aid in the responsible utilization of SDR technology while mitigating public concerns.

With regard to the policy implementation timeline, it is recommended to give priority to policies that swiftly address pressing public concerns, such as enhancing the safety and security of SDRs and ensuring their harmonious coexistence with pedestrians. Following this, policies that support local businesses and job security should be implemented in a phased manner. This staggered approach will contribute to maintaining social stability while effectively fostering the development and application of SDR technology.

In summary, the topic modeling results and the policy recommendations of this study offer valuable insights and concrete strategies for understanding and facilitating the social integration of SDR technology. Through these efforts, we can look forward to a more positive and harmonious role for SDRs in the future urban landscape.

Data collected from comments on social media platforms offer wider insights that provide ground truth and complement survey questionnaires which can gather more detailed perspectives on people's attitudes towards SDRs in certain aspects, as well as information about the respondents. In spite of the valuable insights provided by our study, several limitations need to be addressed in future research. Firstly, our approach to categorizing emotions is rather simplistic. As discussed in our Caveats section, a more nuanced set of emotional labels would allow for a finer distinction between different comments. However, this is contingent upon the size of the dataset. Our dataset is decent but still insufficient to support a more complex array of sentiment labels. This means that under some emotional categories, there may only be a handful or a few dozen entries, leading to an imbalanced dataset. Future studies could benefit from larger and more diverse datasets, which would not only enhance the performance of models but also enable the use of a more sophisticated emotional taxonomy to accurately reflect the nuances in public sentiment. Secondly, the large amount of YouTube comments we collected reflected negative attitudes towards SDRs, but that doesn't mean it will be similar in other social media outlets. Future research could benefit from data collection on other social media platforms like Twitter for a comparative understanding of public sentiment. Lastly, regarding feature engineering and modeling, we only used TF-IDF and N-gram to represent the texts and utilized fairly simple models.

Future work can explore other forms of text representation, such as Word2vec, or more advanced models to achieve better results.

**Conflict of Interest Statement**

**Acknowledgment**

We would like to thank Ruoling Fan for his assistance with data annotation. This research is partially supported by an internal grant from RDE (Realizing the Digital Enterprise) research group at Purdue Polytechnic Institute.

**References**

[1] E. Macioszek, "First and last mile delivery–problems and issues," in *Advanced Solutions of Transport Systems for Growing Mobility: 14th Scientific and Technical Conference" Transport Systems. Theory & Practice 2017" Selected Papers.* Springer, 2018, pp. 147–154.

[2] T. Bosona, "Urban freight last mile logistics—challenges and opportunities to improve sustainability: A literature review," *Sustainability*, vol. 12, no. 21, p. 8769, 2020.

[3] C. Archetti and L. Bertazzi, "Recent challenges in routing and inventory routing: E-commerce and last-mile delivery," *Networks*, vol. 77, no. 2, pp. 255–268, 2021.

[4] H. D. Yoo and S. M. Chankov, "Drone-delivery using autonomous mobility: An innovative approach to future last-mile delivery problems," in *2018 ieee international conference on industrial engineering and engineering management (ieem).* IEEE, 2018, pp. 1216–1220.

[5] A. Comi and L. Savchenko, "Last-mile delivering: Analysis of environment-friendly transport," *Sustainable Cities and Society*, vol. 74, p. 103213, 2021.

[6] D. Jennings and M. Figliozzi, "Study of sidewalk autonomous delivery robots and their potential impacts on freight efficiency and travel," *Transportation Research Record*, vol. 2673, no. 6, pp. 317–326, 2019.

[7] T. Hoffmann and G. Prause, "On the regulatory framework for last-mile delivery robots," *Machines*, vol. 6, no. 3, p. 33, 2018.

[8] D. Weinberg, H. Dwyer, S. E. Fox, and N. Martelaro, "Sharing the sidewalk: Observing delivery robot interactions with pedestrians during a pilot in pittsburgh, pa," *Multimodal Technologies and Interaction*, vol. 7, no. 5, p. 53, 2023.

[9] A. Puig-Pey, J. L. Zamora, B. Amante, J. Moreno, A. Garrell, A. Grau, Y. Bolea, A. Santamaria, and A. Sanfeliu, "Human acceptance in the human-robot interaction scenario for last-mile goods delivery," in *2023 IEEE International Conference on Advanced Robotics and Its Social Impacts (ARSO)*. IEEE, 2023, pp. 33–39.

[10] H. Han, F. M. Li, N. Martelaro, D. Byrne, and S. E. Fox, "The robot in our path: Investigating the perceptions of people with motor disabilities on navigating public space alongside sidewalk robots," in *Proceedings of the 25th International ACM SIGACCESS Conference on Computers and Accessibility*, 2023, pp. 1–6.

[11] C. Bennett, E. Ackerman, B. Fan, J. Bigham, P. Carrington, and S. Fox, "Accessibility and the crowded sidewalk: Micromobility's impact on public space," in *Designing Interactive Systems Conference 2021*, 2021, pp. 365–380.

[12] N. I. Bahari, A. K. Arshad, and Z. Yahya, "Assessing the pedestrians' perception of the sidewalk facilities based on pedestrian travel purpose," in *2013 IEEE 9th International Colloquium on Signal Processing and its Applications*. IEEE, 2013, pp. 27–32.

[13] S. R. Gehrke, C. D. Phair, B. J. Russo, and E. J. Smaglik, "Observed sidewalk autonomous delivery robot interactions with pedestrians and bicyclists," *Transportation research interdisciplinary perspectives*, vol. 18, p. 100789, 2023.

[14] A. Edrisi and H. Ganjipour, "Factors affecting intention and attitude toward sidewalk autonomous delivery robots among online shoppers," *Transportation planning and technology*, vol. 45, no. 7, pp. 588–609, 2022.

[15] A. Pani, S. Mishra, M. Golias, and M. Figliozzi, "Evaluating public acceptance of autonomous delivery robots during covid-19 pandemic," *Transportation research part D: transport and environment*, vol. 89, p. 102600, 2020.

[16] B. Pender, G. Currie, A. Delbosc, and N. Shiwakoti, "Social media use during unplanned transit network disruptions: A review of literature," *Transport reviews*, vol. 34, no. 4, pp. 501–521, 2014.

[17] R. Rahman, K. Redwan Shabab, K. Chandra Roy, M. H. Zaki, and S. Hasan, "Real-time twitter data mining approach to infer user perception toward active mobility," *Transportation research record*, vol. 2675, no. 9, pp. 947–960, 2021.

[18] R. F. Alhujaili and W. M. Yafooz, "Sentiment analysis for youtube videos with user comments: Review," in *2021 International Conference on Artificial Intelligence and Smart Systems (ICAIS)*, 2021, pp. 814–820.

[19] M. Z. Asghar, S. Ahmad, A. Marwat, and F. M. Kundi, "Sentiment analysis on youtube: A brief survey," *arXiv preprint arXiv:1511.09142*, 2015.

[20] A. Krouska, C. Troussas, and M. Virvou, "The effect of preprocessing techniques on twitter sentiment

analysis," in *2016 7th international conference on information, intelligence, systems & applications (IISA)*. IEEE, 2016, pp. 1–5.

[21] A. A. L. Cunha, M. C. Costa, and M. A. C. Pacheco, "Sentiment analysis of youtube video comments using deep neural networks," in *Artificial Intelligence and Soft Computing: 18th International Conference, ICAISC 2019, Zakopane, Poland, June 16–20, 2019, Proceedings, Part I 18*. Springer, 2019, pp. 561–570.

[22] G. Sidorov, S. Miranda-Jiménez, F. Viveros-Jiménez, A. Gelbukh, N. Castro-Sánchez, F. Velásquez, I. Díaz-Rangel, S. Suárez-Guerra, A. Trevino, and J. Gordon, "Empirical study of machine learning based approach for opinion mining in tweets," in *Advances in Artificial Intelligence: 11th Mexican International Conference on Artificial Intelligence, MICAI 2012, San Luis Potosí, Mexico, October 27–November 4, 2012. Revised Selected Papers, Part I 11*. Springer, 2013, pp. 1–14.

[23] T. Mikolov, K. Chen, G. Corrado, and J. Dean, "Efficient estimation of word representations in vector space," *arXiv preprint arXiv:1301.3781*, 2013.

[24] J. Pennington, R. Socher, and C. D. Manning, "Glove: Global vectors for word representation," in *Proceedings of the 2014 conference on empirical methods in natural language processing (EMNLP)*, 2014, pp. 1532–1543.

[25] P. Bojanowski, E. Grave, A. Joulin, and T. Mikolov, "Enriching word vectors with subword information," *Transactions of the association for computational linguistics*, vol. 5, pp. 135–146, 2017.

[26] J. Devlin, M.-W. Chang, K. Lee, and K. Toutanova, "Bert: Pre-training of deep bidirectional transformers for language understanding," *arXiv preprint arXiv:1810.04805*, 2018.

[27] A. N. Muhammad, S. Bukhori, and P. Pandunata, "Sentiment analysis of positive and negative of youtube comments using naïve bayes – support vector machine (nbsvm) classifier," in *2019 International Conference on Computer Science, Information Technology, and Electrical Engineering (ICOMITEE)*, 2019, pp. 199–205.

[28] A.-K. Al-Tamimi, A. Shatnawi, and E. Bani-Issa, "Arabic sentiment analysis of youtube comments," in *2017 IEEE Jordan Conference on Applied Electrical Engineering and Computing Technologies (AEECT)*, 2017, pp. 1–6.

[29] R. Novendri, A. S. Callista, D. N. Pratama, and C. E. Puspita, "Sentiment analysis of youtube movie trailer comments using naïve bayes," *Bulletin of Computer Science and Electrical Engineering*, vol. 1, no. 1, pp. 26–32, 2020.

[30] J. W. Iskandar, Y. Nataliani *et al.*, "Perbandingan naïve bayes, svm, dan k-nn untuk analisis sentimen gadget berbasis aspek," *Jurnal RESTI (Rekayasa Sistem Dan Teknologi Informasi)*, vol. 5, no. 6, pp. 1120–1126, 2021.

[31] R. Ni and H. Cao, "Sentiment analysis based on glove and lstm-gru," in *2020 39th Chinese Control*

*Conference (CCC)*, 2020, pp. 7492–7497.

[32] N. C. Dang, M. N. Moreno-García, and F. De la Prieta, "Sentiment analysis based on deep learning: A comparative study," *Electronics*, vol. 9, no. 3, p. 483, 2020.

[33] S. Styawati, A. Nurkholis, A. A. Aldino, S. Samsugi, E. Suryati, and R. P. Cahyono, "Sentiment analysis on online transportation reviews using word2vec text embedding model feature extraction and support vector machine (svm) algorithm," in *2021 International Seminar on Machine Learning, Optimization, and Data Science (ISMODE)*, 2022, pp. 163–167.

[34] M. Fawzy, M. W. Fakhr, and M. A. Rizka, "Word embeddings and neural network architectures for arabic sentiment analysis," in *2020 16th International Computer Engineering Conference (ICENCO)*, 2020, pp. 92–96.

[35] B. Satya, M. H. S J, M. Rahardi, and F. F. Abdulloh, "Sentiment analysis of review sestyc using support vector machine, naive bayes, and logistic regression algorithm," in *2022 5th International Conference on Information and Communications Technology (ICOIACT)*, 2022, pp. 188–193.

[36] N. B. Epstein and D. H. Baucom, *Enhanced cognitive-behavioral therapy for couples: A contextual approach.*   American Psychological Association, 2002.

[37] J. Suttles and N. Ide, "Distant supervision for emotion classification with discrete binary values," in *International Conference on Intelligent Text Processing and Computational Linguistics*.   Springer, 2013, pp. 121–136.

[38] M. Krommyda, A. Rigos, K. Bouklas, and A. Amditis, "An experimental analysis of data annotation methodologies for emotion detection in short text posted on social media," in *Informatics*, vol. 8, no. 1.   MDPI, 2021, p. 19.

[39] S. Mohammad and F. Bravo-Marquez, "WASSA-2017 shared task on emotion intensity," in *Proceedings of the 8th Workshop on Computational Approaches to Subjectivity, Sentiment and Social Media Analysis.*   Copenhagen, Denmark: Association for Computational Linguistics, Sep. 2017, pp. 34–49. [Online]. Available: https://aclanthology.org/W17-5205

[40] N. I. Tripto and M. E. Ali, "Detecting multilabel sentiment and emotions from bangla youtube comments," in *2018 International Conference on Bangla Speech and Language Processing (ICBSLP)*.   IEEE, 2018, pp. 1–6.

[41] D. M. Blei, A. Y. Ng, and M. I. Jordan, "Latent dirichlet allocation," *Journal of machine Learning research*, vol. 3, no. Jan, pp. 993–1022, 2003.

[42] R. Albalawi, T. H. Yeap, and M. Benyoucef, "Using topic modeling methods for short-text data: A comparative analysis," *Frontiers in artificial intelligence*, vol. 3, p. 42, 2020.

[43] NLTK, "Nltk," 2023, accessed 28/07/2023. [Online]. Available: https://www.nltk.org//

[44] Scikit-learn, "Scikit-learn machine learning in python," 2023, accessed 28/07/2023. [Online]. Available:

https://scikit-learn.org/stable/

[45] H. Face, "Hugging face the ai community building the future." 2023, accessed 28/07/2023. [Online]. Available: https://huggingface.co/

[46] Pytorch, "pytorch," 2023, accessed 28/07/2023. [Online]. Available: https://pytorch.org/

[47] J. Ramos *et al.*, "Using tf-idf to determine word relevance in document queries," in *Proceedings of the first instructional conference on machine learning*, vol. 242, no. 1. Citeseer, 2003, pp. 29–48.

[48] Gensim, "Gensim," 2023, accessed 28/07/2023. [Online]. Available: https://radimrehurek.com/gensim/

[49] M. Röder, A. Both, and A. Hinneburg, "Exploring the space of topic coherence measures," in *Proceedings of the eighth ACM international conference on Web search and data mining*, 2015, pp. 399–408.
34